\documentclass[10pt,twocolumn,letterpaper]{article}

\usepackage{iccv}              

\definecolor{iccvblue}{rgb}{0.21,0.49,0.74}
\usepackage[pagebackref,breaklinks,colorlinks,allcolors=iccvblue]{hyperref}

\def\paperID{512} 
\def\confName{ICCV}
\def\confYear{2025}

\title{NullSwap: Proactive Identity Cloaking Against Deepfake Face Swapping}

\author{Tianyi Wang$^{1}$ \quad Harry Cheng$^{2}$ \quad Xiao Zhang$^{3}$ \quad Yinglong Wang$^{3}$\vspace{0.2em} \\
{\normalsize $^1$Nanyang Technological University} \quad
{\normalsize $^2$Shandong University} \quad \\
{\normalsize $^3$Qilu University of Technology} \quad
}

\begin{document}
\maketitle
\begin{abstract}
Suffering from performance bottlenecks in passively detecting high-quality Deepfake images due to the advancement of generative models, proactive perturbations offer a promising approach to disabling Deepfake manipulations by inserting signals into benign images. However, existing proactive perturbation approaches remain unsatisfactory in several aspects: 1) visual degradation due to direct element-wise addition; 2) limited effectiveness against face swapping manipulation; 3) unavoidable reliance on white- and grey-box settings to involve generative models during training. In this study, we analyze the essence of Deepfake face swapping and argue the necessity of protecting source identities rather than target images, and we propose NullSwap, a novel proactive defense approach that cloaks source image identities and nullifies face swapping under a pure black-box scenario. We design an Identity Extraction module to obtain facial identity features from the source image, while a Perturbation Block is then devised to generate identity-guided perturbations accordingly. Meanwhile, a Feature Block extracts shallow-level image features, which are then fused with the perturbation in the Cloaking Block for image reconstruction. Furthermore, to ensure adaptability across different identity extractors in face swapping algorithms, we propose Dynamic Loss Weighting to adaptively balance identity losses. Experiments demonstrate the outstanding ability of our approach to fool various identity recognition models, outperforming state-of-the-art proactive perturbations in preventing face swapping models from generating images with correct source identities. 
\end{abstract}    
\section{Introduction}
\label{sec:intro}

The flourishing of deep generative models has brought significant development to Deepfake face manipulation techniques~\cite{DeepfakeSurvey[12],JuefeiXu2022[33]}. Due to the improving visual quality of Deepfake images, the domain of passive Deepfake detection has reached bottlenecks in uncovering synthetic artifacts. Therefore, in recent years, researchers have moved on to proactively protect potential victim images in advance. 

\begin{figure}
\centering
\includegraphics[width=\columnwidth]{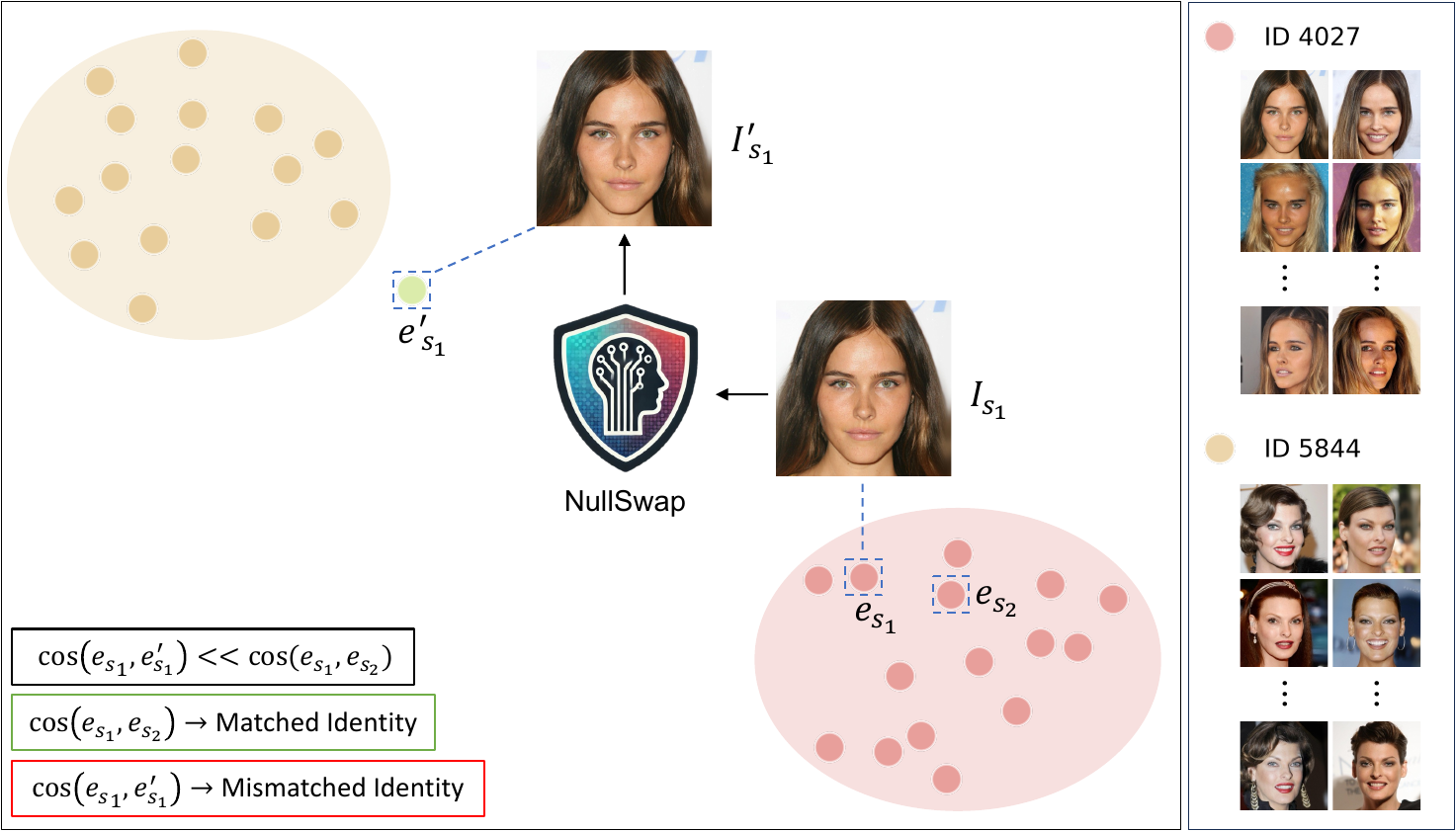}
\caption{ArcFace~\cite{ArcFace[8]} extracted embeddings of images for two facial identities, clustered via t-SNE~\cite{tSNE[17]}. NullSwap aims to insert the perturbation that blinds the identity extractors of face swapping models, such that the perturbed embedding ($e^{'}_{s_1}$) no longer belongs to the original identity of embeddings ($e_{s_1}$ and $e_{s_2}$). }
\label{fig:motivation}
\end{figure}

While traditional passive detection approaches~\cite{DCPT[15],NoiseDF[16],cui2024forensic[34],GenFace[35],FFPP[36],RECCE[37],CADDM[38]} distinguish real and fake after Deepfake manipulations occur, proactive defense~\cite{SepMark[39],IDPMark[14],LampMark[13],huang2021initiative[19],AntiForgery[5]} against Deepfake aims to insert invisible signals into benign images in advance of potential Deepfake manipulations. Proactive perturbations, designed to be imperceptible to human eyes, act as countermoves to distort or nullify Deepfake face manipulations. The perturbations are usually trained to bring significant visual distortions and artifacts to synthetic results regarding certain types of Deepfake manipulations. 

Besides reasonable defensive effects achieved by existing studies, several issues remain unresolved. First, although manipulations such as facial attribute editing and face reenactment have been promisingly handled~\cite{huang2021initiative[19],DisruptingDeepfakes[18],DeepfakeDisrupter[21]}, due to the additional difficulty, nullifying Deepfake face swapping remains largely unattempted and unachieved. Secondly, while most proactive algorithms aim to insert perturbations into the target victim images, we argue the necessity to protect the source victim identities in face swapping, as the swapped identities are the ones in false scandals. Lastly, current solutions mostly rely on practical Deepfake models or imitated surrogate models when learning the perturbation patterns, which brings heavy computing burdens. 

To resolve the aforementioned issues, we propose NullSwap, a novel identity cloaking algorithm designed to proactively defend against malicious Deepfake face swapping attacks. In general, when a celebrity or politician's face is swapped into compromising content, the individuals in the original videos may not require protection. Instead, the focus should be on safeguarding the identities whose faces are swapped, as they are the ones who ultimately face the repercussions of being embroiled in scandals and public controversies. Specifically, since most face swapping algorithms follow the pipeline of obtaining the identity information from the source image and fusing into the target image features for output image reconstruction and production, as demonstrated in Figure~\ref{fig:motivation}, we propose to invisibly cloak the underlying identity information such that incorrect identity embeddings are extracted and the face swapping results are then generated with undesired identities. 

To proactively protect the source identities, for an input image, we first analyze its identity information via an Identity Extraction module. The extracted identity features are then fed to a Perturbation Block to derive the corresponding identity-guided perturbation for identity cloaking. On the other hand, we build a Feature Block to obtai image features from the input image at shallow-depths, pass them to a Cloaking Block along with the learned perturbation, and reconstruct the input image with the perturbation imperceptibly embedded. To ensure generalizability when encountering face swapping algorithms that adopt different identity extractors, we devise a Dynamic Loss Weighting (DLW) mechanism to adaptively adjust the total identity loss based on the values returned by the participating face recognition tools during training. Extensive experiments further demonstrate the ability of NullSwap to cloak the underlying identity information of the original source images and prevent malicious Deepfake face swapping from generating the desired identities even under a black-box scenario such that no generative model participates during training. The contributions of this work are as follows:
\begin{itemize}
\item Upon analyzing the barely discussed essence of face swapping, we propose NullSwap, a novel idea to protect source images and prevent Deepfake face swapping from generating desired identities by embedding perturbations to cloak the identity information. 
\item To our knowledge, we are the first to proactively nullify Deepfake face swapping in a black-box scenario where generative models are omitted in training. Meanwhile, a novel Dynamic Loss Weighting strategy is devised to adaptively balance identity losses for generalizability. 
\item Experiments demonstrate the promising ability of NullSwap to cloak the underlying identity information against face recognition tools and to nullify Deepfake face swapping in deriving undesired identity. 
\end{itemize}
\section{Related Work}
\label{sec:related_work}

\subsection{Deepfake Face Swapping}

Deepfake face swapping performs identity transfer from a source image to the target one. Despite the original deepfakes\footnote{https://github.com/deepfakes/faceswap} and several follow-up improvements~\cite{FaceSwap[31],deepfacelab[29]} that mainly utilize pure Auto-Encoder~\cite{Autoencoder[30]} in an identity-specific manner, the popular face swapping pipeline for arbitrary faces relies on obtaining the facial identity embeddings from the source faces and applying feature injection and fusion when editing the target faces. In 2020, a milestone work, SimSwap~\cite{SimSwap[1]}, extracts the identity embeddings as vectors from the source faces and introduces a novel identity injection module by adopting adaptive instance normalization (AdaIN)~\cite{AdaIN[32]} to integrate features and accomplish target face reconstruction. Similarly, later approaches~\cite{InfoSwap[2],UniFace[3],E4S[4]} follow the same pipeline while improving the visual performance with novel add-ons. For example, besides extracting identity embeddings via ArcFace~\cite{ArcFace[8]}, AP-Swap~\cite{APSwap[28]} provides an efficient global residual attribute-preserving encoder and a landmark-guided feature entanglement module to better preserve the identity-unrelated attributes. 

\begin{figure*}[t!]
\begin{center}
\includegraphics[width=0.95\textwidth]{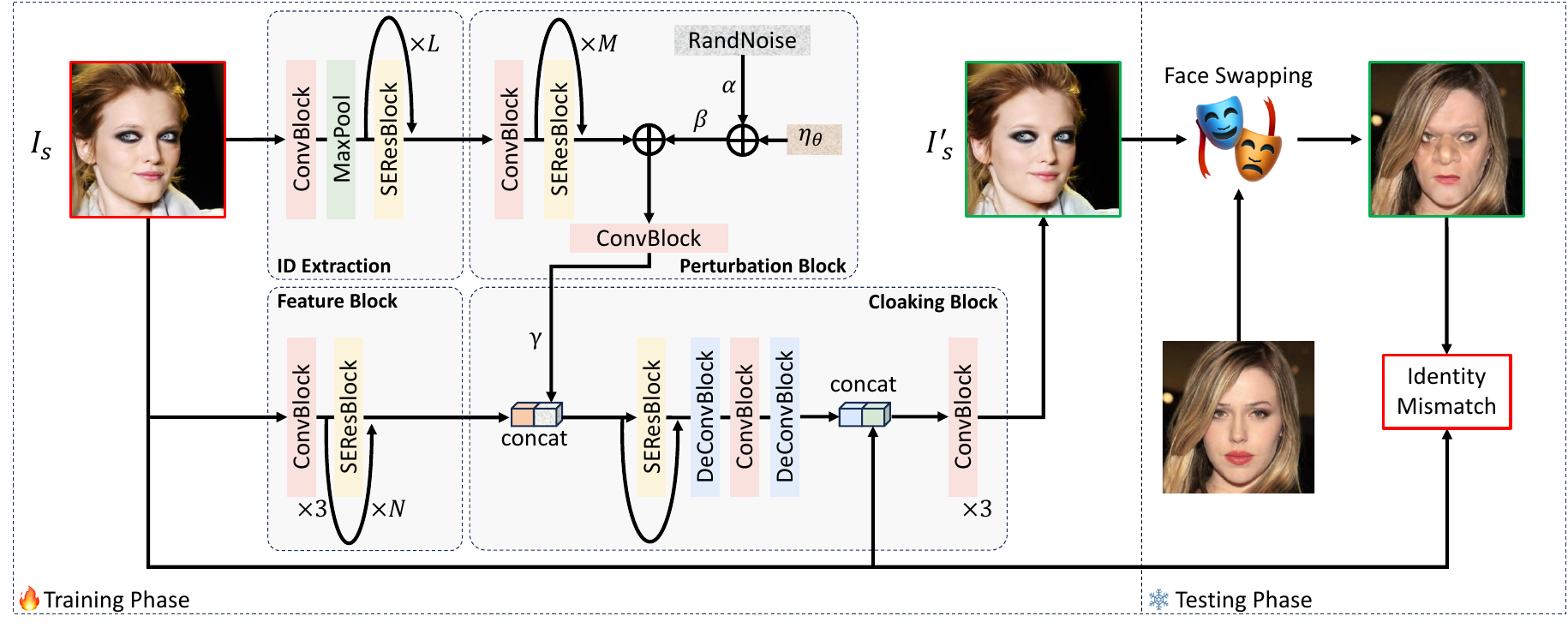}
\end{center}
\caption{Demonstration of the proposed NullSwap framework. The input image $I_s$ is passed through the ID Extraction module and Perturbation Block to generate identity-guided perturbation. A Feature Block executes shallow-level feature extraction on $I_s$ and passes to the Cloaking Block together with the perturbation for the reconstruction of identity-cloaked $I^{'}_s$. }
\label{fig:framework}
\end{figure*}

\subsection{Proactive Perturbation for Deepfake Detection}

Proactive perturbations inserted into the original images are generally aimed to distort Deepfake generation. Most approaches~\cite{Yang2021Defending[20],DeepfakeDisrupter[21],IAP[22],FOUND[24],TAFIM[25]} focus on protecting the victim image that is edited by Deepfake by direct element-wise addition of the perturbation to the image. Ruiz \etal~\cite{DisruptingDeepfakes[18]} proposed a gradient-based method to add imperceptible perturbations to the image and have it distort the Deepfake results. Later, Huang \etal~\cite{huang2021initiative[19]} raised the idea of Initiative defense by building a surrogate model to imitate Deepfake manipulation for devising a poison perturbation generator to obtain the venom for injection. To make it more compatible with human eyes, Wang \etal~\cite{AntiForgery[5]} explored Anti-Forgery to add perturbations to channels of the lab color space. Huang \etal~\cite{CMUA[6]} introduced the Cross-Model Universal Adversarial Watermark (CMUA), a cross-model universal attack pipeline, to attack multiple deepfake models iteratively, and fixed a perturbation generalized to all images and models. Dong \etal~\cite{TCAGAN[23]} proposed TCA-GAN to construct adversarial perturbation for disrupting unknown Deepfake systems via the assistance of a substitute model for face reconstruction. Qu \etal~\cite{DFRAP[7]} designed DF-RAP to provide persistent protection for facial images under OSN compressions. A compression approximation GAN is trained to explicitly model OSN compression and then incorporated as a sub-module of the target Deepfake model. 

However, existing literatures have barely considered the potential visual quality issue caused by direct element-wise addition~\cite{DualDefense[26],Defeating2022[27]}, and the popular white-box and grey-box settings have brought high computational costs during model training. Moreover, considering the actual victim of Deepfake face swapping, the protection of the source image as the identity provider has rarely been discussed. 
\section{Method}

\subsection{Problem Formulation}
\label{sec:problem_form}
Ever since the first occurrence of Deepfake face swapping, various milestone studies~\cite{FFPP[36],SimSwap[1],UniFace[3]} have advanced the research domain with better generative quality, but the general pipeline remains relatively fixed. In specific, for a source face that provides the desired facial identity and a target face whose facial identity is modified while maintaining the rest attributes, facial identity feature extraction usually happens to the former, and feature fusion followed by image reconstruction happens on the latter. In practice, a victim being face-swapped onto indelicate content is mired in scandal because of the high identity similarity as displayed in the synthetic media, somehow regardless of the visual quality. Therefore, although most existing proactive countermoves attempt to distort the generated facial images with different types of perturbations, we argue that, considering the essence of Deepfake face swapping, the goal of proactive defense is to prevent the face swapping models from extracting the correct identity information so that synthetic faces are generated with undesired and unexpected facial identities. In this way, Deepfake face swapping is proactively nullified while the generative models are untouched. 

In this study, we propose NullSwap, a novel idea to achieve identity cloaking by embedding identity-guided perturbations into the source images, such that the perturbed images are visually identical to the original clean ones while the identity embeddings extracted by face recognition tools are vastly shifted (Figure~\ref{fig:motivation}). Since the objective can be achieved with the assistance of multiple face recognition tools, to the best of our knowledge, our proposed approach is the first proactive perturbation approach against face swapping in a black-box scenario without generative models participating during the training phase. 

\subsection{NullSwap Framework}
\label{sec:nullswap_framework}
For an unprotected source image $I_s$ under the risk of malicious Deepfake face swapping, we propose NullSwap (Figure~\ref{fig:framework}) to derive a perturbation protected image $I^{'}_s$. $I_s$ is first passed to the Identity Extraction (ID Extraction) module for identity feature extraction, and then a Perturbation Block to derive the identity-guided perturbation that can cloak the identity information. Meanwhile, a Feature Block is applied to the input image $I_s$ for shallow-level feature extraction. The image feature and perturbation are passed to a Cloaking Block and the reconstructed image $I^{'}_s$ is ultimately derived. 

\noindent\textbf{Identity Extraction.} Unlike face recognition tools~\cite{ArcFace[8],FaceNet[9],VGGFace[10]} that provide identity embeddings in vector format, we establish an ID Extraction module to analyze the input image and maintain identity features that are dominant toward our objectives in matrix format. In specific, a ConvBlock, consisting of a CNN layer, a batch normalization, and a ReLU activation, followed by a max-pooling operation, is applied to $I_s$. To moderately maintain critical features from early stages and efficiently analyze the inter-channel correlation, we apply $L$ consecutive SEResBlocks, where each block adopts the ResNet~\cite{ResNet[40]} bottleneck block and the squeeze-and-excitation networks (SENet)~\cite{SENet[41]}. The feature sufficiency in this module provides the training capacity to consistently fool multiple face recognition tools. 

\noindent\textbf{Perturbation Block.} The extracted identity features are fed to a Perturbation Block to derive identity-guided perturbation for the ultimate identity cloaking. Specifically, following a ConvBlock for feature refinement, $M$ SEResBlocks are sequentially applied for hierarchical feature aggregation, preserving the identity features that are most relevant for perturbation and suppressing the less relevant ones. Meanwhile, considering the possible overfitting to specific identity patterns, we introduce a degree of randomness to the perturbation. In particular, after initializing a random noise drawn from the normal distribution, we derive the adaptive random noise following
\begin{equation}
\textrm{RandNoise} = \beta \cdot (\alpha \cdot \textrm{RandNoise} + \eta),
\label{eq:rand_noise}
\end{equation}
where $\eta$ is a learnable noise, and $\alpha$ and $\beta$ are learnable parameters to adaptively adjust the magnitude of randomness. The random noise is then added to the deterministic perturbation, as exhibited in Figure~\ref{fig:framework}.

\noindent\textbf{Feature Block.} Considering the visual flaws caused by direct perturbation addition in existing work, we propose to embed the learned perturbation into latent image features. Therefore, for the input image $I_s$, we build a Feature Block to perform shallow-level feature extraction. Specifically, after three consecutive ConvBlocks for local feature analysis and dimension adjustment, we apply $N$ SEResBlocks to enhance feature adaptability and context awareness before image reconstruction for $I^{'}_s$.

\noindent\textbf{Cloaking Block.} In this Cloaking Block, we reconstruct the input image while invisibly embedding the perturbation. First, by assigning a learnable weight $\gamma$ to the perturbation, we concatenate it with the shallow-level image feature and apply a feature-level reconstruction via a SEResBlock, a DeConvBlock, a ConvBlock, and a DeConvBlock to balance perturbation embedding and visual fidelity. To elaborate, the DeConvBlock consists of the upsampling operation, batch normalization, and ReLU activation. In general, this feature-level reconstruction gradually restores spatial structures while effectively integrating the identity-guided perturbation. 

In the end, for the feature-level reconstructed result containing a first-stage fusion of the perturbation and image features, we concatenate the input image to it and apply a final image-level reconstruction via three sequential ConvBlocks to derive $I^{'}_s$. With the two reconstructions at different dimension levels, we ensure that the reconstructed image remains perceptually similar to the original with mere artifacts while successfully cloaking the source identity.

\subsection{Dynamic Loss Weighting}

In this study, the main objective is to fool the face recognition tools upon identity embedding extraction. Since the adopted embedding extractor varies in different face swapping algorithms, we introduce multiple face recognition tools to ensure the generalizability of our approach. We design Dynamic Loss Weighting (DLW) to adaptively balance loss objectives drawn via different face recognition algorithms during training. It consists of two core components to determine the relative importance of each loss at each time state, namely, loss variance and relative progress. Loss variance measures the stability of each loss over a recent training window. Specifically, losses with higher variance, indicating instability or noise, are assigned lower weights to ensure stable optimization. On the other hand, relative progress evaluates the rate of improvement for each loss. In particular, losses showing slower progress are prioritized, allowing them to catch up and ensuring balanced contributions across tasks.

By introducing DLW, the identity loss is denoted as
\begin{equation}
    \mathcal{L}_\textrm{id}(t_e, t_b) = \Sigma_{i=1}^c\hat{w_i}(t_e, t_b) \cdot \mathcal{L}_i(t_b),
\label{eq:dlw}
\end{equation}
where for each $\mathcal{L}_i$ of the total of $c$ loss functions at iteration $t_b$ of epoch $t_e$, the weight $\hat{w_i}$ is determined by
\begin{equation}
    \hat{w_i}(t_e, t_b) = \frac{c \cdot w_i(t_e, t_b)}{\Sigma_{j=1}^c w_j(t_e, t_b)},
\end{equation}
which denotes the normalization process that aligns each weight $w_i$ regarding the magnitude of different losses. Before normalization, the weight $w_i$ is derived by
\begin{equation}
    w_i(t_e, t_b) = \max\left(
    \frac{1}{\max(\alpha \cdot \sigma_i^2(t_b) + \beta(t_e) \cdot (1 + \Delta_i(t_b)), \epsilon_d)}, \epsilon_w
    \right),
\end{equation}
where $\alpha$ is a constant regularization factor that balances the loss variance and relative progress. $\beta$ relies on the current epoch $t_e$ by
\begin{equation}
    \beta(t_e) = \min\left(\beta_{\textrm{init}} + \gamma \cdot \min(t_e, t_e^*), \beta^*\right),
\end{equation}
where $t_e^*$ and $\beta^*$ are the upper limit values for $t_e$ and $\beta$, respectively. The variance $\sigma^2_i$ is computed based on the recent $k$ consecutive loss values with respect to the current iteration $t_b$ following
\begin{equation}
    \sigma^2_i(t_b)=\frac{1}{k}\sum_{j=0}^{k-1}\left(\mathcal{L}_{i}(t_b-j)-\frac{1}{k}\Sigma^{k-1}_{m=0}(\mathcal{L}_i(t_b - m))\right)^2,
\end{equation}
and $\Delta_i$ is the relative progress computed based on loss values of the current and the previous iterations
\begin{equation}
    \Delta_i = \max\left(\frac{\mathcal{L}_{i}(t_b-1) - \mathcal{L}_i(t_b)}{\mathcal{L}_i(t_b-1) + \epsilon_p}, -1\right).
\end{equation}

DLW ensures that losses contributing more stable and consistent gradients maintain their influence, while noisy or under-performing losses are appropriately adjusted. By combining loss variance and relative progress adaptively, DLW aligns the training process with the overarching goal of optimizing all objectives fairly. Furthermore, DLW is computationally efficient and directly operates on loss values, dynamically adjusting loss contributions without requiring additional parameters or gradient-level manipulation.

\subsection{Objective Functions}

In this study, the main goal is to maintain image reconstruction quality while promisingly nullifying face recognition tools. On the one hand, considering the pixel-wise similarity, we employ the Mean Squared Error (MSE) loss denoted as 
\begin{equation}
\mathcal{L_\textrm{MSE}} = \lVert I_s - I^{'}_s \rVert_2.
\end{equation}
On the other hand, regarding the potential artifacts that are negligible by MSE loss such as blurring and loss of high-frequency details, we adopt the Learned Perceptual Image Patch Similarity (LPIPS) loss~\cite{LPIPS[42]} to measure perceptual similarity based on deep feature representations following
\begin{equation}
\mathcal{L_\textrm{LPIPS}} = \Sigma_l \lVert \phi_l(I_s) - \phi_l(I^{'}_s) \rVert_2^2,
\end{equation}
where $\phi_l$ represents the feature activation from a pre-trained AlexNet~\cite{AlexNet[43]} at layer $l$. 

Furthermore, we train a discriminator from scratch along with the NullSwap model training, where the discriminator acts as a binary classifier with $N$ ConvBlocks. The objective function of the discriminator $D$ is denoted as
\begin{equation}
\mathcal{L}_\textrm{dis} = -\mathbb{E}(\log D(I_s)) +\mathbb{E}\log(1-D(I^{'}_s)).
\end{equation}
While the discriminator is iteratively trained, it provides guidance on the visual quality of perturbed images following
\begin{equation}
\mathcal{L}_D = -\mathbb{E}(\log D(I^{'}_s)).
\end{equation}

As for the performance in cloaking the identity information, to guarantee generalizability, we adopt multiple face recognition tools to provide the cosine similarities between identity embeddings $e_s$ and $e^{'}_s$ extracted from $I_s$ and $I^{'}_s$, respectively. The formula is denoted as
\begin{equation}
\mathcal{L}_i(e_s, e^{'}_s) = \frac{e_s \cdot e^{'}_s}{\lVert e_s \rVert \lVert e^{'}_s \rVert},
\end{equation}
and the identity loss is computed following DLW as demonstrated in Eqn.(\ref{eq:dlw}).

The total loss is computed as 
\begin{equation}
\mathcal{L}_\textrm{total} = \lambda_{\textrm{id}}\mathcal{L}_{\textrm{id}} + \lambda_{\textrm{MSE}}\mathcal{L}_{\textrm{MSE}} + \lambda_{\textrm{LPIPS}}\mathcal{L}_{\textrm{LPIPS}} + \lambda_D\mathcal{L}_D,
\label{eq:total_loss}
\end{equation}
with pre-defined coefficients $\lambda_{\textrm{id}}$, $\lambda_{\textrm{MSE}}$, $\lambda_{\textrm{LPIPS}}$, and $\lambda_D$.
\section{Experiments}

\subsection{Implementation Details}

In the experiment, we adopted the popular human face dataset, CelebA-HQ~\cite{CelebAHQ[44]}, which contains 30,000 high-quality facial images with 6,217 unique identities, following the official split for training, validation, and testing. Meanwhile, we adopted another classical face dataset, Labeled Faces in the Wild (LFW)~\cite{LFW[45]} with 5,749 facial identities, for the cross-dataset validation purpose. Regarding the constant coefficients for the number of blocks in Figure~\ref{fig:framework}, we set $L=4$, $M=3$, and $N=5$. We employed ArcFace~\cite{ArcFace[8]} and FaceNet~\cite{FaceNet[9]} as the assisting face recognition tools during training for the DLW computation and set $\alpha=3$, $\beta_\textrm{init}=0.5$, $\beta^*=2$, $\gamma=0.1$, $k=30$, and $\epsilon_d = \epsilon_w = 1e-6$. For the coefficients in Eqn.(\ref{eq:total_loss}), we set $\lambda_{\textrm{id}}=0.08$, $\lambda_{\textrm{MSE}}=1.8$, $\lambda_{\textrm{LPIPS}}=1.2$, and $\lambda_D=0.1$ for total loss. During training, the learning rates for NullSwap and the discriminator are $5e-4$ and $1e-4$, respectively, for $60$ training epochs. Experiments are conducted on images resized to $256 \times 256$. Our model is implemented using PyTorch on 8 Tesla A100 GPUs with a batch size of 256. 

\begin{table}[t!]
\centering
\begin{tabular}{lccc}
\toprule
Models & PSNR$\uparrow$ & SSIM$\uparrow$ & LPIPS$\downarrow$ \\
\midrule
Initiative~\cite{huang2021initiative[19]} & 39.3803 & 0.9544 & 0.0200 \\
Anti-Forgery~\cite{AntiForgery[5]} & 38.0705 & 0.9530 & 0.0282 \\
CMUA~\cite{CMUA[6]} & 38.6395 & 0.9504 & 0.0333 \\
DF-RAP~\cite{DFRAP[7]} & 38.8466 & 0.9349 & 0.0511 \\
NullSwap (Ours) & \textbf{41.3097} & \textbf{0.9864} & \textbf{0.0049} \\
\bottomrule
\end{tabular}
\caption{Visual quality evaluation of perturbed images. Best performance marked in \textbf{bold}. }
\label{tab:visual_quality}
\end{table}

\begin{figure}[t!]
\centering
\includegraphics[width=\columnwidth]{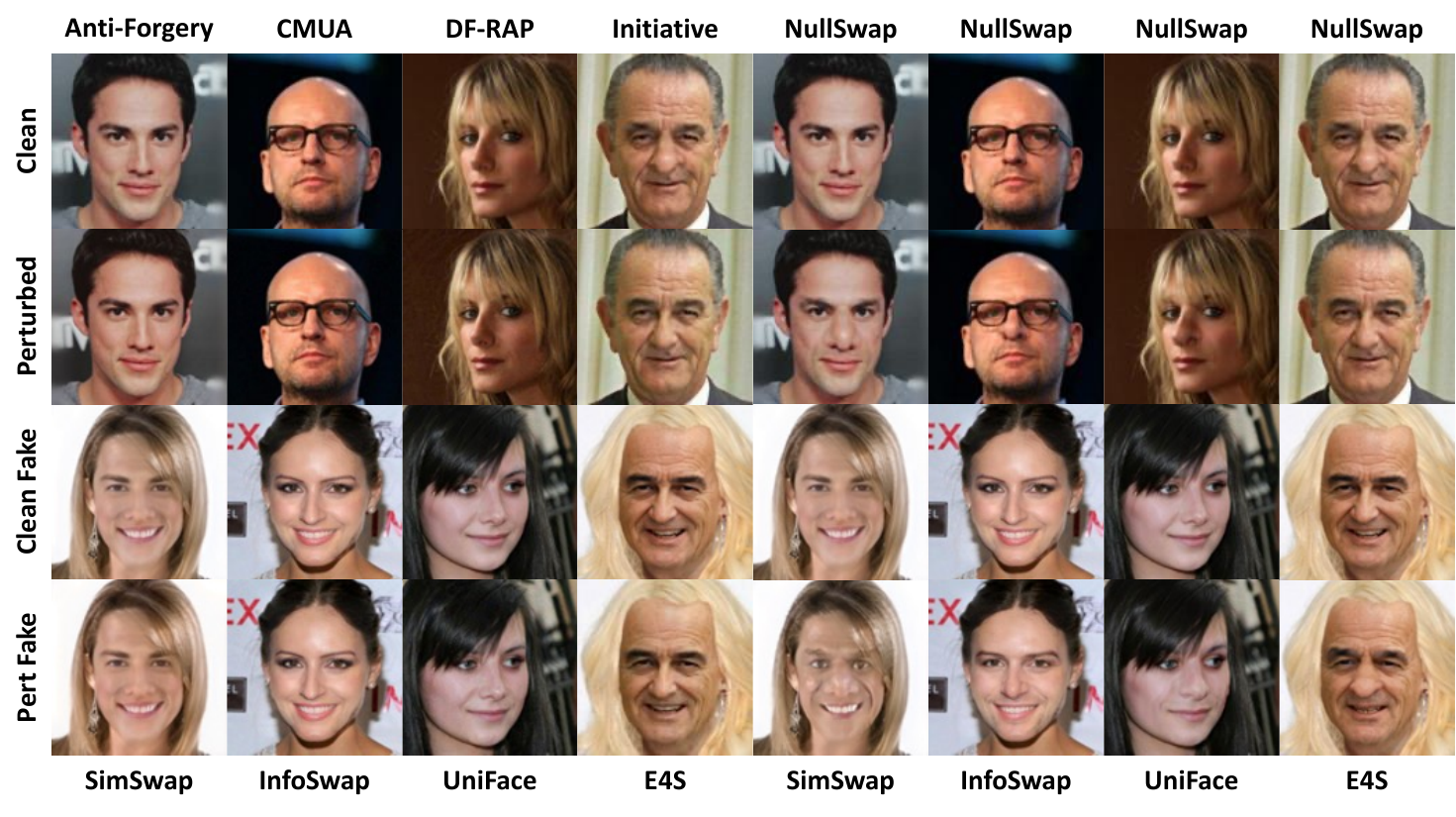}
\caption{Visualization of face swapping performance regarding different perturbation algorithms against different generative models. }
\label{fig:visualization}
\end{figure}

\subsection{Evaluation Results}

\begin{table*}[t!]
\centering
\resizebox{\textwidth}{!}{
\begin{tabular}{lcccccccccccc}
\toprule
Models & \multicolumn{2}{c}{Clean} & \multicolumn{2}{c}{Initiative~\cite{huang2021initiative[19]}} & \multicolumn{2}{c}{Anti-Forgery~\cite{AntiForgery[5]}} & \multicolumn{2}{c}{CMUA~\cite{CMUA[6]}} & \multicolumn{2}{c}{DF-RAP~\cite{DFRAP[7]}} & \multicolumn{2}{c}{NullSwap (Ours)} \\
\midrule
Type & & & \multicolumn{2}{c}{Grey-Box} & \multicolumn{2}{c}{Grey-Box} & \multicolumn{2}{c}{White-Box} & \multicolumn{2}{c}{Grey-Box} & \multicolumn{2}{c}{Black-Box} \\
\midrule
 & Acc@5 & Acc@1 & Acc@5$\downarrow$ & Acc@1$\downarrow$ & Acc@5$\downarrow$ & Acc@1$\downarrow$ & Acc@5$\downarrow$ & Acc@1$\downarrow$ & Acc@5$\downarrow$ & Acc@1$\downarrow$ & Acc@5$\downarrow$ & Acc@1$\downarrow$ \\
\midrule
ArcFace~\cite{ArcFace[8]} & 0.980 & 0.976 & 0.974 & 0.968 & 0.977 & 0.975 & 0.979 & 0.976 & 0.977 & 0.974 & \textbf{0.771} & \textbf{0.628} \\
FaceNet~\cite{FaceNet[9]} & 0.940 & 0.918 & 0.952 & 0.925 & 0.921 & 0.862 & 0.920 & 0.865 & 0.955 & 0.920 & \textbf{0.740} & \textbf{0.590} \\
VGGFace~\cite{VGGFace[10]} & 0.910 & 0.853 & 0.917 & 0.856 & 0.919 & 0.858 & 0.925 & 0.864 & 0.909 & 0.847 & \textbf{0.674} & \textbf{0.529} \\
SFace~\cite{SFace[11]} & 0.804 & 0.791 & 0.794 & 0.720 & 0.800 & 0.732 & 0.799 & 0.720 & 0.779 & 0.680 & \textbf{0.658} & \textbf{0.513} \\
\midrule
Average & 0.909 & 0.885 & 0.909 & 0.867 & 0.904 & 0.857 & 0.906 & 0.856 & 0.905 & 0.855 & \textbf{0.711} & \textbf{0.565} \\
\bottomrule
\end{tabular}
}
\caption{Top-5 and top-1 accuracies regarding the identity matching performance after applying perturbation algorithms. Lower values refer to better perturbation performance. Best performance marked in \textbf{bold}. }
\label{tab:id_pert_performance}
\end{table*}

In this section, we validated the performance of the proposed NullSwap model. First, we evaluated the visual quality after embedding perturbations and discussed the performance numerically and visually. Then, we checked the identity cloaking performance of the embedded perturbation compared to state-of-the-art proactive perturbation algorithms by adopting their published best model weights. After that, ablation studies and discussions are conducted. 

\subsubsection{Visual Quality.} 

As reported in Table~\ref{tab:visual_quality}, we computed the average peak signal-to-noise ratio (PSNR), structural similarity index measure (SSIM), and the learned perceptual image patch similarity (LPIPS) regarding the clean and perturbed images. Specifically, considering that Initiative~\cite{huang2021initiative[19]}, Anti-Forgery~\cite{AntiForgery[5]}, CMUA~\cite{CMUA[6]}, and DF-RAP~\cite{DFRAP[7]} all insert perturbations by direct element-wise addition, unavoidable visual artifacts can be perceptible to both human eyes and deep neural networks. As a result, our approach achieves state-of-the-art visual performance, with PSNR above 40 while the rest are below, SSIM above 0.98 while the rest hover around 0.95, and LPIPS below 0.005 while the rest are above 0.020. 

We also visualized some samples in Figure~\ref{fig:visualization} such that the first and second rows display the clean and perturbed images via different proactive perturbation approaches. It can be observed that, although achieving a PSNR above 38 and an SSIM around 0.95 guarantees acceptable visual qualities for the perturbed images, those with worse LPIPS values demonstrate artifacts in terms of illumination and blurring. 

\subsubsection{Identity Cloaking Performance.} 
\label{sec:id_cloak_performance}

\noindent\textbf{Identity Cloaking on Source Faces.} After deriving the perturbed image $I^{'}_s$, we tested whether $I^{'}_s$ still remains within the same identity cluster or is successfully cloaked. To do so, a face recognition task evaluation is executed on each perturbed image regarding the identity label provided by CelebA-HQ. In particular, we computed the top-5 and top-1 accuracies for the cosine similarities between identity embeddings and that of each $I^{'}_s$. To prove the generalizability of our proposed approach, besides including ArcFace~\cite{ArcFace[8]} and FaceNet~\cite{FaceNet[9]} that participate during the training phase, we also adopted the well-trained VGGFace~\cite{VGGFace[10]} and SFace~\cite{SFace[11]} as face recognition tools for NullSwap to fool during the testing phase. 

As listed in Table~\ref{tab:id_pert_performance}, although values vary depending on the optimal performance of different face recognition tools, the top-5 and top-1 accuracies on clean testing images generally maintain the highest performance as expected. On the other hand, while the state-of-the-art comparative proactive perturbation models have caused accuracy degradations in most entries, the differences from the clean setting are mostly below 0.1 in absolute values except DF-RAP when facing SFace. Moreover, each of the comparative models even unexpectedly encounters greater top-5 and top-1 accuracies than that in the clean setting. This might be caused by the cases where perturbations fail to achieve the desired goal but somehow advance the feature extraction efficacy upon insertion. As a result, our proposed NullSwap achieves the average top-5 and top-1 face recognition accuracies of 0.711 and 0.565, respectively, promisingly outperforming state-of-the-art proactive perturbation algorithms such that we fooled each face recognition tool the best in the competition. 

\begin{table*}[t!]
\centering
\resizebox{\textwidth}{!}{
\begin{tabular}{lcccccccccc}
\toprule
Models & \multicolumn{2}{c}{Initiative~\cite{huang2021initiative[19]}} & \multicolumn{2}{c}{Anti-Forgery~\cite{AntiForgery[5]}} & \multicolumn{2}{c}{CMUA~\cite{CMUA[6]}} & \multicolumn{2}{c}{DF-RAP~\cite{DFRAP[7]}} & \multicolumn{2}{c}{NullSwap (Ours)} \\
\midrule
 & ArcFace$\downarrow$ & VGGFace$\downarrow$ & ArcFace$\downarrow$ & VGGFace$\downarrow$ & ArcFace$\downarrow$ & VGGFace$\downarrow$ & ArcFace$\downarrow$ & VGGFace$\downarrow$ & ArcFace$\downarrow$ & VGGFace$\downarrow$ \\
\midrule
SimSwap~\cite{SimSwap[1]} & 0.928 & 0.897 & 0.924 & 0.892 & 0.921 & 0.930 & 0.468 & 0.431 & \textbf{0.217} & \textbf{0.240} \\
InfoSwap~\cite{InfoSwap[2]} & 0.941 & 0.919 & 0.903 & 0.922 & 0.947 & 0.888 & 0.913 & 0.849 & \textbf{0.375} & \textbf{0.359} \\
UniFace~\cite{UniFace[3]} & 0.987 & 0.960 & 0.987 & 0.968 & 0.983 & 0.965 & 0.947 & 0.891 & \textbf{0.369} & \textbf{0.329} \\
E4S~\cite{E4S[4]} & 0.925 & 0.893 & 0.924 & 0.901 & 0.920 & 0.900 & 0.880 & 0.855 & \textbf{0.398} & \textbf{0.368} \\
DiffSwap~\cite{DiffSwap[55]} & 0.660 & 0.657 & 0.663 & 0.655 & 0.658 & 0.648 & 0.636 & 0.631 & \textbf{0.310} & \textbf{0.352} \\
\midrule
Average & 0.888 & 0.865 & 0.880 & 0.868 & 0.886 & 0.866 & 0.769 & 0.731 & \textbf{0.334} & \textbf{0.330} \\
\bottomrule
\end{tabular}
}
\caption{Identity similarity evaluation of face swapping results on perturbed images compared to that on the clean ones using various Deepfake models via perturbation algorithms on CelebA-HQ. Best performance marked in \textbf{bold}. }
\label{tab:swap_pert_performance}
\end{table*}

\begin{table*}[t!]
\centering
\resizebox{\textwidth}{!}{
\begin{tabular}{lcccccccccc}
\toprule
Models & \multicolumn{2}{c}{Clean} & \multicolumn{2}{c}{Initiative~\cite{huang2021initiative[19]}} & \multicolumn{2}{c}{CMUA~\cite{CMUA[6]}} & \multicolumn{2}{c}{DF-RAP~\cite{DFRAP[7]}} & \multicolumn{2}{c}{NullSwap (Ours)} \\
\midrule
 & Acc@5 & Acc@1 & Acc@5$\downarrow$ & Acc@1$\downarrow$ & Acc@5$\downarrow$ & Acc@1$\downarrow$ & Acc@5$\downarrow$ & Acc@1$\downarrow$ & Acc@5$\downarrow$ & Acc@1$\downarrow$ \\
\midrule
ArcFace~\cite{ArcFace[8]} & 0.923 & 0.910 & 0.864 & 0.801 & 0.925 & 0.903 & 0.891 & 0.841 & \textbf{0.627} & \textbf{0.491} \\
FaceNet~\cite{FaceNet[9]} & 0.822 & 0.708 & 0.726 & 0.555 & 0.822 & 0.702 & 0.773 & 0.621 & \textbf{0.324} & \textbf{0.210} \\
VGGFace~\cite{VGGFace[10]} & 0.704 & 0.550 & 0.588 & 0.429 & 0.701 & 0.555 & 0.648 & 0.494 & \textbf{0.294} & \textbf{0.186} \\
SFace~\cite{SFace[11]} & 0.553 & 0.445 & 0.485 & 0.369 & 0.565 & 0.455 & 0.542 & 0.437 & \textbf{0.348} & \textbf{0.225} \\
\midrule
Average & 0.751 & 0.653 & 0.666 & 0.539 & 0.753 & 0.654 & 0.714 & 0.598 & \textbf{0.398} & \textbf{0.278} \\
\midrule
PSNR$\uparrow$ & \multicolumn{2}{c}{--} &\multicolumn{2}{c}{34.1381} & \multicolumn{2}{c}{37.8710} & \multicolumn{2}{c}{30.6551} & \multicolumn{2}{c}{\textbf{42.8132}} \\
SSIM$\uparrow$ & \multicolumn{2}{c}{--} &\multicolumn{2}{c}{0.8380} & \multicolumn{2}{c}{0.9392} & \multicolumn{2}{c}{0.8824} & \multicolumn{2}{c}{\textbf{0.9882}} \\
LPIPS$\downarrow$ & \multicolumn{2}{c}{--} &\multicolumn{2}{c}{0.3102} & \multicolumn{2}{c}{0.0517} & \multicolumn{2}{c}{0.1422} & \multicolumn{2}{c}{\textbf{0.0049}} \\
\bottomrule
\end{tabular}
}
\caption{Top-5 and top-1 accuracies regarding the identity matching performance after applying perturbation algorithms on LFW. Lower values refer to better perturbation performance. Best performance marked in \textbf{bold}. Anti-Forgery~\cite{AntiForgery[5]} is omitted since LFW does not contain the attribute information it requires. }
\label{tab:id_pert_performance_lfw}
\end{table*}

\noindent\textbf{Identity Cloaking on Face Swapping.} Moreover, considering the essential goal of NullSwap is to cloak the identity information and prevent Deepfake face swapping from generating images with desired identities, we tested the model performance by introducing various state-of-the-art popular face swapping algorithms, namely, SimSwap~\cite{SimSwap[1]}, InfoSwap~\cite{InfoSwap[2]}, UniFace~\cite{UniFace[3]}, E4S~\cite{E4S[4]}, and DiffSwap~\cite{DiffSwap[55]}. Note that the face swapping models only appear in the testing phase in our black-box setting. In this experiment, we adopted the two face recognition tools, ArcFace~\cite{ArcFace[8]} and VGGFace~\cite{VGGFace[10]}, proved to have the best performance and stability\footnote{https://github.com/serengil/deepface}, as evaluators. Specifically, to evaluate the strength of perturbations in affecting the face swapping results, we first generated face swapping results via each generative model using clean images. Then, we repeated the same pipeline along with the identical source-target pairs using the perturbed images derived by each proactive perturbation model. For face swapping results based on perturbed faces, we compute the cosine similarity with respect to the corresponding clean face swapping results based on their identity embeddings. 

As reported in Table~\ref{tab:swap_pert_performance}, Initiative~\cite{huang2021initiative[19]}, Anti-Forgery~\cite{AntiForgery[5]} and CMUA~\cite{CMUA[6]} are observed to derive face swapping results that have cosine similarities all around 0.9, except DiffSwap~\cite{DiffSwap[55]}, regarding the clean ones, regardless of specific face swapping algorithms. This implies that, although demonstrating reasonable defensive ability against easier tasks such as attribute editing, as reported in the original work, the inserted perturbations generally fail to distort or nullify Deepfake face swapping. On the other hand, DF-RAP~\cite{DFRAP[7]}, being evaluated against SimSwap in the original work, appears to be effective in nullifying SimSwap~\cite{SimSwap[1]}, with low cosine similarities of 0.468 and 0.431 for ArcFace and VGGFace as expected. However, the perturbations are observed to lack generalizability toward other face swapping models, leading to similarities around or above 0.850. DiffSwap~\cite{DiffSwap[55]}, a diffusion-based face swapping method, although executes unstable generation results due to its original performance, is still highly nullified by our approach, deriving identity similarities as low as 0.310 and 0.352 for ArcFace and VGGFace, respectively. As a result, our proposed method consistently nullifies different state-of-the-art face swapping models by cloaking the identity information, achieving average similarities of 0.334 and 0.330 for ArcFace and VGGFace, respectively, with individual similarities all below 0.4. 

This is consistent with the visualized results in Figure~\ref{fig:visualization}. Particularly, the perturbed fake faces are highly identical to the clean fake faces in the columns of Anti-Forgery, CMUA, and DF-RAP, and no visible distortion can be observed. As for Initiative, only lighting artifacts can be observed. Contrarily, in the first two columns of our proposed NullSwap, obvious visual differences are observed between the clean and perturbed face swapping results. For the rest two columns, detailed but critical identity-specific differences are caused by the identity cloak on the nose, mouth, and cheek areas. 

\noindent\textbf{Cross-Dataset Validation.} To further validate the generalizability of our proposed NullSwap on unseen data, we conducted cross-dataset validations on the Labeled Faces in the Wild (LFW)~\cite{LFW[45]} dataset. Experiments are conducted following the same pipeline as Table~\ref{tab:id_pert_performance} except Anti-Forgery is omitted due to unprovided attribute information in LFW. As Table~\ref{tab:id_pert_performance_lfw} shows, while CMUA~\cite{CMUA[6]} generally maintains similar top-5 and top-1 identity recognition accuracies as the clean ones for all four face recognition tools, Initiative~\cite{huang2021initiative[19]} and DF-RAP~\cite{DFRAP[7]} achieve the defensive goal at certain levels but fail to preserve reasonable visual qualities. On the other hand, our proposed identity cloaking strategy leads to the largest degradations with 0.398 and 0.278 on average for top-5 and top-1 accuracies, while consistently maintaining promising image visual quality. In conclusion, NullSwap demonstrates outstanding generalizability on unseen datasets such as LFW. 

\begin{table}[t!]
\centering
\resizebox{\columnwidth}{!}{
\begin{tabular}{lcccc}
\toprule
Models & Session 1 & Session 2 & Session 3 & DLW (Ours) \\
\midrule
ArcFace$\downarrow$ & 0.653 & 0.843 & 0.846 & 0.628 \\
FaceNet$\downarrow$ & 0.758 & 0.546 & 0.613 & 0.590 \\
VGGFace$\downarrow$ & 0.680 & 0.504 & 0.600 & 0.529 \\
SFace$\downarrow$ & 0.576 & 0.495 & 0.525 & 0.513 \\
\midrule
Average & 0.667 & 0.597 & 0.646 & 0.565 \\
\midrule
PSNR$\uparrow$ & 42.1461 & 40.3650 & 40.8079 & 41.3097 \\
SSIM$\uparrow$ & 0.9872 & 0.9814 & 0.9747 & 0.9864 \\
LPIPS$\downarrow$ & 0.0042 & 0.0075 & 0.0063 & 0.0049 \\
\bottomrule
\end{tabular}
}
\caption{Ablation study on the strategies regarding identity losses. The NullSwap model is trained with Session 1 (ArcFace loss), Session 2 (FaceNet loss), Session 3 (average of ArcFace and FaceNet losses), and the proposed Dynamic Loss Weighting (DLW).}
\label{tab:ablation}
\end{table}

\subsubsection{Ablation Study}
\label{sec:ablation}

In this section, we validated the performance of our proposed Dynamic Loss Weighting (DLW) strategy regarding the case when introducing multiple face recognition tools to help ensure generalizability. We conducted four training sessions in different settings with respect to the participating face recognition tools in this study. Specifically, Session 1 adopts ArcFace~\cite{ArcFace[8]} solely to provide cosine similarity as the identity loss, and similarly, Session 2 adopts FaceNet~\cite{FaceNet[9]} solely. As for Session 3, both ArcFace and FaceNet are adopted during training, and the identity loss is computed by averaging their provided cosine similarities. Lastly, we kept our optimal solution, DLW, as Session 4. 

Since the only difference happens in the identity loss, the visual quality of all sessions does not heavily fluctuate. Therefore, we focused on analyzing the identity cloaking performance of different sessions. Following the same evaluation process as in Table~\ref{tab:id_pert_performance}, the top-1 face recognition accuracy of each training session is exhibited in Table~\ref{tab:ablation}. Session 1, although successfully decreases the accuracy on ArcFace, comes up with the highest accuracies on FaceNet, VGGFace, and SFace, leading to the largest average accuracy of 0.667. On the other hand, despite maintaining high accuracy for ArcFace, Session 2 trained with FaceNet is able to drag the accuracies of the rest down to lower levels. This might be due to the feature overlapping of the way they derive identity embeddings. However, training by simply averaging the loss for ArcFace and FaceNet in Session 3 fails to achieve better performance. Contrarily, our proposed Dynamic Loss Weighting performs consistently for all face recognition tools. This proves the fact that identity losses of ArcFace and FaceNet suffer fluctuations between batches during the training session, which critically affects the achievement of the objectives. Furthermore, DLW considers history losses, which not only stabilizes the fluctuations among batches but also adaptively adjusts weight updating regarding loss values that are stuck at certain points. As a result, although not achieving the optimal results for all entries, our optimal session with DLW achieves the best average performance on identity cloaking at 0.565. 

\subsubsection{Discussion}

Deepfake threats have happened in recent years in cases such as false scandals and economic fraud~\cite{DeepfakeSurvey[12]} relying on the success of reconstructing the source face identity. While non-professionals usually have tolerance for the visual quality upon mere glances but tend to believe the truthfulness once recognize the facial identities, nullifying Deepfake face swapping and making the generated images with undesired identities can effectively resolve the threats. In this study, despite the original accuracies of face recognition tools can vary and the performance of face swapping algorithms may be unstable, our proposed method consistently decreases the similarities between the cloaked identity information and the clean ones while maintaining acceptable visual quality, as demonstrated in Tables~\ref{tab:visual_quality} to~\ref{tab:id_pert_performance_lfw}. Besides, due to the diversity in existing and future Deepfake face swapping algorithms, the white- and grey-box settings with generative models participating during training can progressively become computation consuming while still lacking generalizability. As proved in Sections~\ref{sec:id_cloak_performance} and~\ref{sec:ablation}, introducing DLW for multiple identity recognition tools consistently achieves promising performance across seen and unseen face recognition tools and face swapping models. 

\section{Conclusion}

In this study, we analyze the essence of the Deepfake face swapping attack and claim the necessity of protecting the source images that provide facial identities, the real victims, rather than the target images that are visually edited. We propose an identity cloaking proactive approach, NullSwap, to adaptively embed identity-guided perturbations into source images for nullifying Deepfake face swapping such that synthetic faces are generated with undesired facial identities. In addition to preserving reasonable image visual quality, our proposed approach demonstrates an outstanding ability to fool various identity extractors of different Deepfake face swapping algorithms in cross-dataset and cross-model scenarios. Besides outperforming state-of-the-art proactive perturbations, the novel claim of protecting the source identities rather than target images provides innovative insights in the research domain, and the black-box scenario without generative models participating during training deserves further deep analyses in future studies.

{
    \small
    \bibliographystyle{ieeenat_fullname}
    \bibliography{main}
}

\end{document}